\documentclass[nohyperref]{article}

\usepackage{microtype}
\usepackage{graphicx}
\usepackage{booktabs} 

\usepackage{hyperref}

\usepackage[accepted]{wfvml2022}

\usepackage{amsmath}
\usepackage{amssymb}
\usepackage{mathtools}
\usepackage{amsthm}

\usepackage[capitalize,noabbrev]{cleveref}

\theoremstyle{plain}

\theoremstyle{definition}

\theoremstyle{remark}

\usepackage[inline]{enumitem}

\usepackage{caption}
\usepackage{subcaption}
\usepackage{wrapfig}
\usepackage{xcolor}

\usepackage{xfrac}

\newcommand{\up}[1]{\overline{#1}}
\newcommand{\lo}[1]{\underline{#1}}

\let\vec\mathbf

\DeclarePairedDelimiter\abs{\lvert}{\rvert}%
\DeclarePairedDelimiter\norm{\lVert}{\rVert}%

\makeatletter
\let\oldabs\abs
\def\abs{\@ifstar{\oldabs}{\oldabs*}}
\let\oldnorm\norm
\def\norm{\@ifstar{\oldnorm}{\oldnorm*}}
\makeatother

\DeclareMathOperator{\R}{\mathbb{R}}

\DeclareMathOperator{\relu}{ReLU}

\newcommand\nnfun[1]{f_{\mathcal{#1}}}

\usepackage[textsize=tiny]{todonotes}

\wfvmltitlerunning{Optimized Symbolic Interval Propagation for Neural Network Verification}

\begin{document}

\twocolumn[
\wfvmltitle{Optimized Symbolic Interval Propagation for Neural Network Verification}



\wfvmlsetsymbol{equal}{*}

\begin{wfvmlauthorlist}
\wfvmlauthor{Philipp Kern}{yyy}
\wfvmlauthor{Marko Kleine Büning}{yyy}
\wfvmlauthor{Carsten Sinz}{yyy}
\end{wfvmlauthorlist}

\wfvmlaffiliation{yyy}{Department of Computer Science, Karlsruhe Institute of Technology, Karlsruhe, Germany}

\wfvmlcorrespondingauthor{Philipp Kern}{philipp.kern@kit.edu}
\wfvmlcorrespondingauthor{Marko Kleine Büning}{marko.kleinebuening@kit.edu}
\wfvmlcorrespondingauthor{Carsten Sinz}{carsten.sinz@kit.edu}

\wfvmlkeywords{Machine Learning, Formal Verification, Symbolic Interval Propagation}

\vskip 0.3in
]



\printAffiliationsAndNotice{}  

\begin{abstract}
Neural networks are increasingly applied in safety critical domains,
their verification thus is gaining importance.
A large class of recent algorithms for proving input-output relations of
feed-forward neural networks are based on linear relaxations and
symbolic interval propagation. However, due to variable dependencies, the
approximations deteriorate with increasing depth of the network.
In this paper we present \textsc{DPNeurifyFV}, a novel branch-and-bound solver for $\relu$ networks with low dimensional input-space that is based on symbolic interval propagation with fresh variables and input-splitting.
A new heuristic for choosing the fresh variables allows to ameliorate the dependency problem,
while our novel splitting heuristic, in combination with several other improvements, speeds up the branch-and-bound procedure.
We evaluate our approach on the airborne collision avoidance networks ACAS Xu 
and demonstrate 
runtime improvements compared to state-of-the-art tools.
\end{abstract}

\section{Introduction}
\label{sec:introduction}

Neural networks (NNs) are considered state-of-the-art solutions for many machine learning tasks and are increasingly applied in a variety of different areas. However, the complex structure of neurons, layers, and weights often renders their behavior incomprehensible to humans.
This can have a severe impact when NNs are applied to safety-critical systems such as airborne collision avoidance \cite{julian2016policy} or self-driving cars \cite{Bojarski2016}.
The increased application of NNs in security-critical areas necessitates formal guarantees about their behavior and functionality. 

In general, the goal is to verify input-output relations of NNs. 
For this purpose, there are a number of approaches based on SMT \cite{Ehlers2017,huang2017safety} or MILP \cite{Katz2017,katz2019marabou} solving as well as on abstract interpretation theory \cite{muller2021scaling,muller2021prima,singh2019beyond,tran2020nnv,xiang2018output} including symbolic interval propagation (SIP) \cite{fischer2019dl2,gehr2018ai2,Henriksen20,Wang18a}.

Na\"{i}ve interval propagation is a fast analysis method to propagate lower and upper bounds through a network. 
However, it suffers from the \textit{dependency problem} and thus can lead to very coarse overapproximations.
Although this problem can be weakened by the application of relational domains like zonotopes, error-based symbolic interval propagation or backward substitution of symbolic intervals \cite{Wang18b, Singh19}, these solutions are either restricted to parallel relaxations of the activation functions or require a quadratic number of backsubstitution passes.
While not as effective in reducing the dependency problem, the introduction of fresh variables in SIP \cite{Paulsen20} requires only a single forward pass and allows for the use of non-parallel relaxations.

In this paper, we present an approach that verifies input-output relations of feed-forward $\relu$-networks with low-dimensional input space via input splitting and optimized symbolic interval propagation with fresh variables.

The main contributions of this work are: 
(i) We re-examine and optimize the introduction of fresh variables in symbolic interval propagation by presenting heuristics for the amount of variables to introduce per layer and which neurons to prioritize.
(ii) We present a novel input splitting heuristic taking into account internal overapproximation (similar to \cite{Henriksen21} for error-based symbolic interval propagation and neuron splitting)
(iii) We improve counterexample generation by utilizing information of the symbolic bounds instead of just sampling the center of the input hyperrectangle.

\section{Foundations and Related Work }
\label{sec:foundations}

\subsection{Neural Networks.}
\label{ssec:neural_networks}

A feed-forward neural network $\mathcal{N}$ represents a function $f_{\mathcal{N}} : \R^{d_0} \rightarrow \R^{d_L}$ and consists of an input layer, multiple hidden layers and an output layer \cite{Goodfellow2016}.
The $l$-th layer ($l \geq 1$) comprises $d_l$ neurons $\vec{n}_l$, whose activation values are calculated as an affine function of the values of the previous layer's neurons (or the inputs, for $l = 0$, identifying $\vec{n}_0$ with the input vector $\vec{x}$), and then fed through a non-linear activation function $\sigma$ according to
	$\vec{n}_l = \sigma\left(W_l \vec{n}_{l-1} + \vec{b}_l\right)$,
where $W_l$ is the weight matrix and $\vec{b}_l$ is the bias vector of layer $l$.
Vector $\vec{y} = \vec{n}_L$ is the output of the NN.
In this work, we only consider cases, where $\sigma$ is the ReLU activation function defined as $\relu(x) = \max(0, x)$ and the weights of the NN are fixed.

\subsection{Neural Network Verification.}
\label{ssec:nn_verification}

The field of NN verification is concerned with verifying that a NN computing a function $\vec{y} = f_{\mathcal{N}}(\vec{x})$ always satisfies a property $P(\vec{y})$ for all inputs 
in a set $\mathcal{X} \subseteq \mathbb{R}^{d_0}$.
In most of the current literature, $P$ is represented as a set of linear constraints on the output of the NN, while the input is constrained to values within a hyperrectangle 
    $\mathcal{H} = \{\vec{x} \in \mathbb{R}^{d_0} ~|~ \lo{x}_i \leq x_i \leq \up{x}_i ~\text{for all}~ 0 \leq i < d_0 \}$ given by lower and upper bounds, $\lo{x}_i$
    and $\up{x}_i$, for each dimension.

If $P$ consists of only a single inequality $\vec{c}^T\vec{y} + b \leq 0$, the problem of NN verification can be rephrased as an optimization problem \cite{Bunel20}:
\begin{align}
	\max_{\vec{x}} ~ & \vec{c}^T\vec{y} + b\\
			   s.t.~ & \vec{y} = f_{\mathcal{N}}(\vec{x})\\
			  	     & \lo{x}_i \leq x_i \leq \up{x}_i, ~\forall i \enspace.
	\label{eq:nn-opt}
\end{align}

If the maximal value $v^*$ of the optimization problem is larger than $0$, the property is violated and 	a concrete input vector $\vec{x}^*$ with 
	$\vec{c}^Tf_{\mathcal{N}}(\vec{x}^*) + b > 0$
	can be computed as a counterexample to the validity of property $P$.
If we can prove $v^* \leq 0$ or we can find some upper bound $v^*_\text{ub}$
with $v^* \leq v^*_\text{ub} \leq 0$, then the property is guaranteed to hold for
network $\mathcal{N}$ and all $x \in \mathcal{H}$.

This formulation can be generalized to properties $P$ consisting of multiple inequalities.
We always assume that these are given in the form $\vec{c}^T\vec{y} + b \leq 0$. Thus,
by maximizing the maximum violation for all inequality constraints, we can check whether $P$ holds or not.
Due to the piecewise-linear $\relu$ activations, NN verification 
may require case splitting, and is in general NP-complete \cite{Katz2017}.
Still many approaches have been proposed, for example based on mixed integer linear programming \cite{Cheng2017,Tjeng2017}, SMT solvers \cite{Ehlers2017,Katz2017}, or
abstract interpretation based on different set representations
\cite{Bak20,Henriksen20,Singh18,Singh19,Wang18b,Wang18a,Zhang18}.

\begin{figure*}
	\begin{center}
	\includegraphics[width=0.85\linewidth]{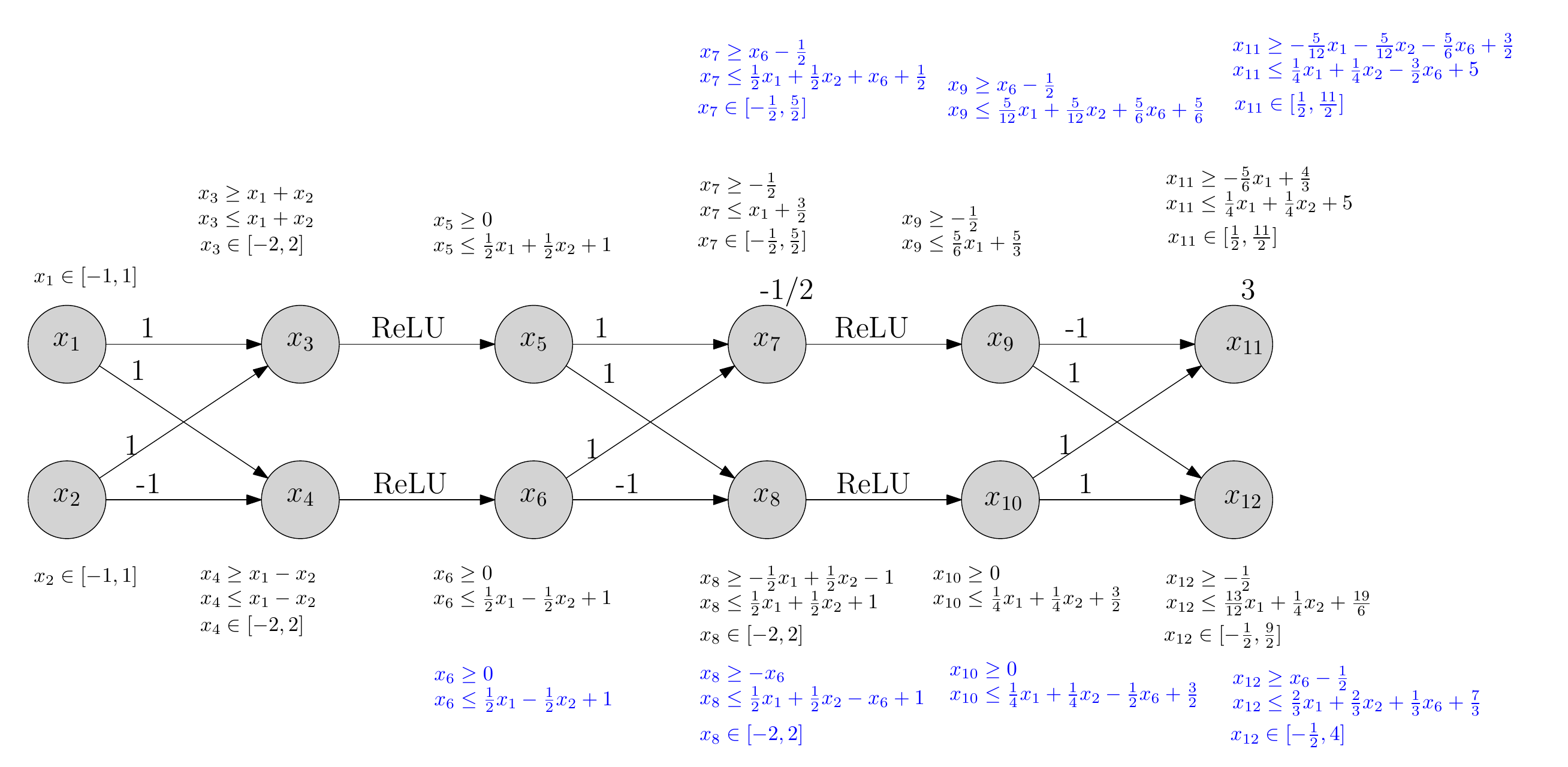}
	\caption{Example for forward propagation of symbolic intervals. Weights are written as edge labels, the bias, if non-zero, is indicated above to the right of the nodes. Symbolic interval propagation is shown in black, while the effects of introducing a fresh variable $x_6$ for the constraints are shown in blue, resulting in an improved bound for $x_{12}$.}
	\label{fig:symbolic_intervals}
	\end{center}
\end{figure*}

\subsection{Symbolic Interval Propagation}
\label{ssec:sip}

In abstract interpretation, a suitable set representation $S_0$ encompassing the admissible values for the input set $\mathcal{X}$ is propagated through the layers of the NN until an output set $S_L$ is obtained, which overapproximates the true output set $\{\vec{y} = \nnfun{N}(\vec{x}) ~|~ \vec{x} \in \mathcal{X}\}$ of the NN.
Despite the NP-completeness of NN verification, abstract interpretation approaches allow for efficient verification, if one can easily check whether $P(\vec{y})$ is satisfied for all $\vec{y} \in S_L$. 

In the following paragraphs we want to give an overview over symbolic intervals \cite{Wang18b,Wang18a}, which are one of the possible set representations in abstract interpretation for NNs.
In particular, we want to highlight two main factors that affect the quality
that can be obtained from symbolic interval propagation: The dependency problem and the overapproximation introduced, when symbolic intervals are propagated through the non-linear activation functions.

In traditional interval arithmetic \cite{Moore1966}, intervals $[\lo{x}_i, \up{x}_i]$ with concrete lower and upper bounds $\lo{x}_i \leq x_i \leq \up{x}_i$
are chosen as enclosing set representations for the true value of $x_i$.
To obtain a valid enclosing interval for operations like addition, say $x + y$, where $x \in [\lo{x}, \up{x}], y \in [\lo{y}, \up{y}]$, the interval extension $\widehat{+}$ of the addition operation is defined as $[\lo{x}, \up{x}] \widehat{+} [\lo{y}, \up{y}] = [\lo{x} + \lo{y}, \up{x} + \up{y}]$.
A similar extension can be defined for subtraction as $[\lo{x}, \up{x}] \widehat{-} [\lo{y}, \up{y}] = [\lo{x} - \up{y}, \up{x} - \lo{y}]$.
Although interval arithmetic is cheap, most of the time it computes very conservative overapproximations.
If we subtract the interval $[\lo{x}, \up{x}]$ representing $x \in \mathbb{R}$ from itself, we obtain $[\lo{x}, \up{x}] - [\lo{x}, \up{x}] = [\lo{x} - \up{x}, \up{x} - \lo{x}] $ instead of the expected interval $[0, 0]$.
The reason for this is the \emph{dependency problem}.
Information about dependencies between intervals is lost during computation,
and different operands representing the same variable can vary independently within their lower and upper bounds.
This problem is amplified by chained computations, where the outputs of one stage serve as 
inputs for the next stage \cite{Stolfi1997}. 
Therefore abstract interpretation for (deep) NNs is particularly affected by the dependency problem.

As a countermeasure, Wang et al. \cite{Wang18b,Wang18a} introduce symbolic intervals $[lb(\vec{x}), ub(\vec{x})]$, where possible values of quantities $z \in \mathbb{R}$ are enclosed by pairs of linear functions $lb(\vec{x}) = \vec{l}^T\vec{x} + l_0$ and $ub(\vec{x}) = \vec{u}^T\vec{x} + u_0$ in the input variables $\vec{x}$ that $z$ depends upon, instead of concrete bounds. This allows to capture more dependency information.
Note that each input variable $x_i$ only occurs once in each bound, which precludes it from attaining its concrete lower and upper bounds simultaneously.
For vector-valued quantities $\vec{z} \in \mathbb{R}^n$, $\vec{lb}(\vec{x})$ and $\vec{ub}(\vec{x})$ are affine mappings, where the $\vec{l}^T$ and $\vec{u}^T$ form the rows of a coefficient matrix and $\vec{l}_0$ and $\vec{u}_0$ are vectors.

Propagation of a symbolic interval vector $[\vec{lb}(\vec{x}), \vec{ub}(\vec{x})]$ through affine mappings $\vec{y} = W\vec{x} + \vec{b}$ with constant $W$ and $\vec{b}$ results in another symbolic interval 
defined by
	$\vec{lb'}(\vec{x}) = W^+ \vec{lb}(\vec{x}) + W^- \vec{ub}(\vec{x}) + \vec{b}$ and
	$\vec{ub'}(\vec{x}) = W^+ \vec{ub}(\vec{x}) + W^- \vec{lb}(\vec{x}) + \vec{b}$,
where $W^+$ and $W^-$ denote the positive and the negative entries of $W$ respectively. 
As only linear operations are used, no overapproximation is incurred by this propagation \cite{Wang18a}.

Occurrences of the same input variable with opposite signs can be combined,
such that there is only one occurrence of the input variable in each of the output's symbolic lower and upper bounds, which is set to either its lower or upper bound upon concretization.
To illustrate this behaviour, the value of $x_7$ in Fig.~\ref{fig:symbolic_intervals} is calculated as $x_7 = x_5 + x_6 - \frac{1}{2}$, where the true values are enclosed by symbolic intervals $x_5 \in [0, \frac{1}{2} ~ x_1 + \frac{1}{2} ~ x_2 + 1]$ and $x_6 \in [0, \frac{1}{2} ~ x_1 - \frac{1}{2} ~ x_2 + 1]$ for $x_1, x_2 \in [-1, 1]$.
We obtain the symbolic upper bound of $x_7$ as
\begin{align*}
	ub_7(& x_1, x_2) = ub_5(x_1, x_2) + ub_6(x_1, x_2) - \frac{1}{2} \\
				   &= \left(\frac{1}{2} ~ x_1 + \frac{1}{2} ~ x_2 + 1\right) + \left(\frac{1}{2} ~ x_1 - \frac{1}{2} ~ x_2 + 1\right) - \frac{1}{2} \\
				   &= x_1 + \frac{3}{2} \text{ ,}
\end{align*}
where the opposite signs of $x_2$ even cancel completely.
The concrete upper bound of $\up{x}_7 = \frac{5}{2}$ can be obtained by setting $x_1 \in [-1, 1]$ to its upper bound since it occurs with positive coefficient in $ub_7(x_1, x_2)$.
If we had just used the concrete interval enclosure of $x_5 \in [\lo{x}_5, \up{x}_5] = [0, 2], x_6 \in [\lo{x}_6, \up{x}_6] = [0, 2]$, we would have gotten an unnecessarily loose enclosing interval of $x_7 \in [-\frac{1}{2}, \frac{7}{2}]$, as we would have implicitly used the upper bound $\up{x}_2$ in $\up{x}_5$ and the lower bound $\lo{x}_2$ in $\up{x}_6$.

In order to propagate symbolic intervals through the piecewise linear activation function $\relu(x)$, for $x \in [lb(\vec{x}), ub(\vec{x})]$, the symbolic interval is concretized to obtain a concrete interval enclosure $x \in [l, u]$.
The symbolic interval enclosing the output of $y = \relu(x)$ is then computed as 
\begin{align*}
lb'(\vec{x}) = \begin{cases}
					0 &, u \leq 0 \\
					\lo{\relu}(lb(\vec{x})) &, l < 0 < u \\
					lb(\vec{x}) &, l \geq 0
			   \end{cases}
\end{align*}
and 
\begin{align*}
ub'(\vec{x}) = \begin{cases}
					0 &, u \leq 0 \\
					\up{\relu}(ub(\vec{x})) &, l < 0 < u \\
					ub(\vec{x}) &, l \geq 0
			   \end{cases} \text{ ,}
\end{align*}
where the $\relu$ function is linearly overapproximated by lower and upper relaxations $\lo{\relu}(x)$ and $\up{\relu}(x)$, it is unstable or crossing ($l < 0 < u$), while no overapproximation is necessary if it is fixed-active ($l \geq 0$) or fixed-zero ($u \leq 0$).

While error-based symbolic interval propagation \cite{Henriksen20} and zonotope propagation \cite{Singh18} are restricted to parallel $\relu$ relaxations, the methods of \cite{Zhang18} and \cite{Singh19} use different slopes for the upper relaxation
	$\up{\relu}(x) = \frac{u}{u - l} x - \frac{ul}{u - l}$
and the lower relaxation
\begin{align*}
	\lo{\relu}(x) = \begin{cases}
						0, ~ u < |l| \\
						x, ~ u \geq |l|
					\end{cases}
\end{align*}
where $\lo{\relu}(x)$ is chosen adaptively to reduce the area of the introduced overapproximation.

This $\relu$ relaxation is also used for analysis of the example NN shown in Fig.~\ref{fig:symbolic_intervals}.
Utilizing the fact that $\relu(x)$ is monotonic in $x$, the authors of \cite{Wang18b} introduce, what we call separate relaxations.
They calculate separate concrete intervals $lb(\vec{x}) \in [l_l, l_u]$ and $ub(\vec{x}) \in [u_l, u_u]$.
The symbolic bounds $lb(\vec{x})$ and $ub(\vec{x})$ are then separately propagated through the lower parallel relaxation of $\relu(x)$ for $x \in [l_l, l_u]$ and the upper parallel relaxation of $\relu(x)$ for $x \in [u_l, u_u]$.

Although symbolic interval propagation ensures that \emph{input variables} never attain their lower and upper bound at the same time (assuming they are not equal), when concretizing the symbolic interval, no such guarantees can be given for intermediate neurons. For example, we implicitly use the \emph{upper} relaxation of neuron $x_6$ in network in Fig.~\ref{fig:symbolic_intervals} to calculate the symbolic upper bounds $ub_7(\vec{x})$ of $x_7$, which we use for the calculation of the $ub_9(\vec{x})$, which is in turn used for $ub_{12}(\vec{x})$.
This symbolic upper bound of $x_{12}$, however is calculated using $ub_{10}(\vec{x})$, which relies on $ub_8(\vec{x})$, which is calculated using the \emph{lower} relaxation of $x_6$.
 
Zonotope- \cite{Singh18} and error-based symbolic interval propagation \cite{Henriksen20} as well as backsubstitution methods \cite{Zhang18, Singh19} can solve this problem.
Both methods, however, cannot utilize separate relaxations and while the former methods are restricted to only parallel relaxations, the number of backsubstitution passes required by the latter methods grows quadratically in the number of layers.

In the context of verification of equivalence for NNs, the authors of \cite{Paulsen20} introduce symbolic interval propagation with the additional introduction of fresh variables that are used for some of the overapproximated neurons.
In order to introduce a fresh variable for a neuron $x_{n+1}$ contained within the symbolic interval $[lb_{n+1}(\vec{x}), ub_{n+1}(\vec{x})]$ depending on the input $\vec{x} \in \mathbb{R}^n$ of the NN, they store its symbolic bounds for later substitution and instead propagate the new symbolic interval $[x_{n+1}, x_{n+1}]$, ensuring that $x_{n+1}$ can never simultaneously operate at its lower and upper symbolic bounds.
A symbolic interval $[lb'(\vec{x}), ub'(\vec{x})]$ only depending on the inputs of the NN, can be recovered from a symbolic interval $[l_0 + \sum_{i=1}^{n+k} l_i x_i, u_0 + \sum_{i=1}^{n+k} u_i x_i]$ with $k$ fresh variables according to
\begin{align*}
	lb'(\vec{x}) &= l_0 + \sum_{i=1}^n l_i x_i + \sum_{i = n+1}^{n+k} \left( l_i^+ lb_i(\vec{x}) + l_i^- ub_i(\vec{x}) \right) \\
	ub'(\vec{x}) &= u_0 + \sum_{i=1}^n u_i x_i + \sum_{i = n+1}^{n+k} \left( u_i^+ ub_i(\vec{x}) +  u_i^- lb_i(\vec{x}) \right)
\end{align*}
where the $l_i^+ = \max(0, l_i), ~ u_i^+ = \max(0, u_i)$ represent positive and the $l_i^- = \min(0, l_i), ~ u_i^- = \min(0, u_i)$ represent negative coefficients in the lower and upper symbolic bounds respectively.
Some dependency information is lost, however, as only symbolic bounds with respect to the input variables are stored in order to avoid expensive chained backsubstitutions.
Furthermore, the introduction of fresh variables in later layers impedes the cancellation of earlier variables (which would again increase the dependency problem), because all of their coefficients are set to zero in the new symbolic interval for the new fresh variable.
Nevertheless, this approach allows for the use of non-parallel $\relu$ relaxations and is capable of decreasing the impact of the dependency issue, when fresh variables are introduced for fitting unstable neurons.

For example, introducing a fresh variable for neuron $x_6$ in Fig.~\ref{fig:symbolic_intervals}, results in an improvement in the concrete upper bound of the output $x_{12}$.
The corresponding symbolic intervals that result from the introduction of $x_6$ as a fresh variable are shown in blue.

\begin{figure*}
\centering
	\begin{subfigure}{.45\textwidth}
		\centering
		\includegraphics[width=0.9\linewidth]{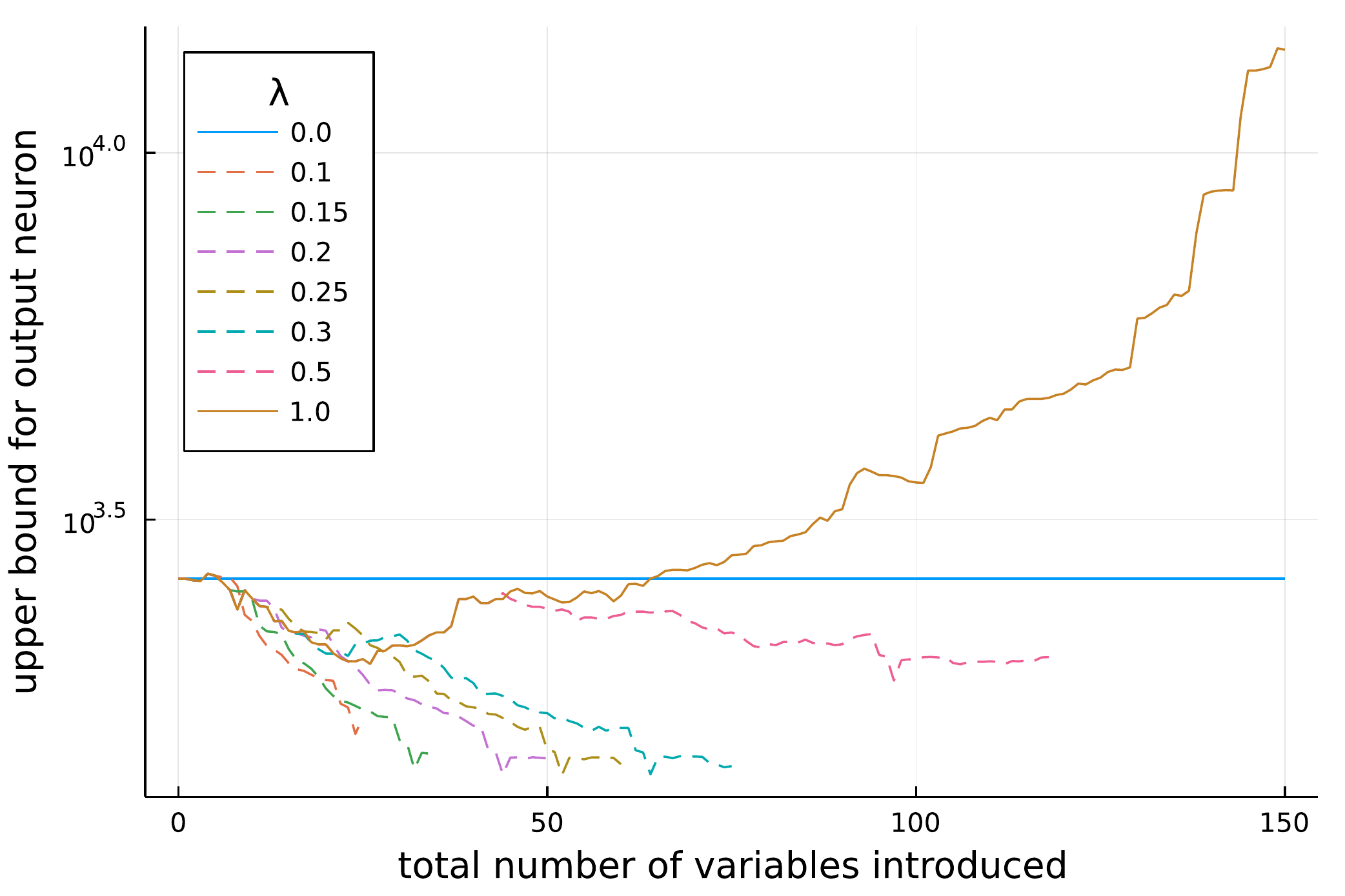}
	\end{subfigure}
	\begin{subfigure}{.45\textwidth}
		\centering
		\includegraphics[width=0.9\linewidth]{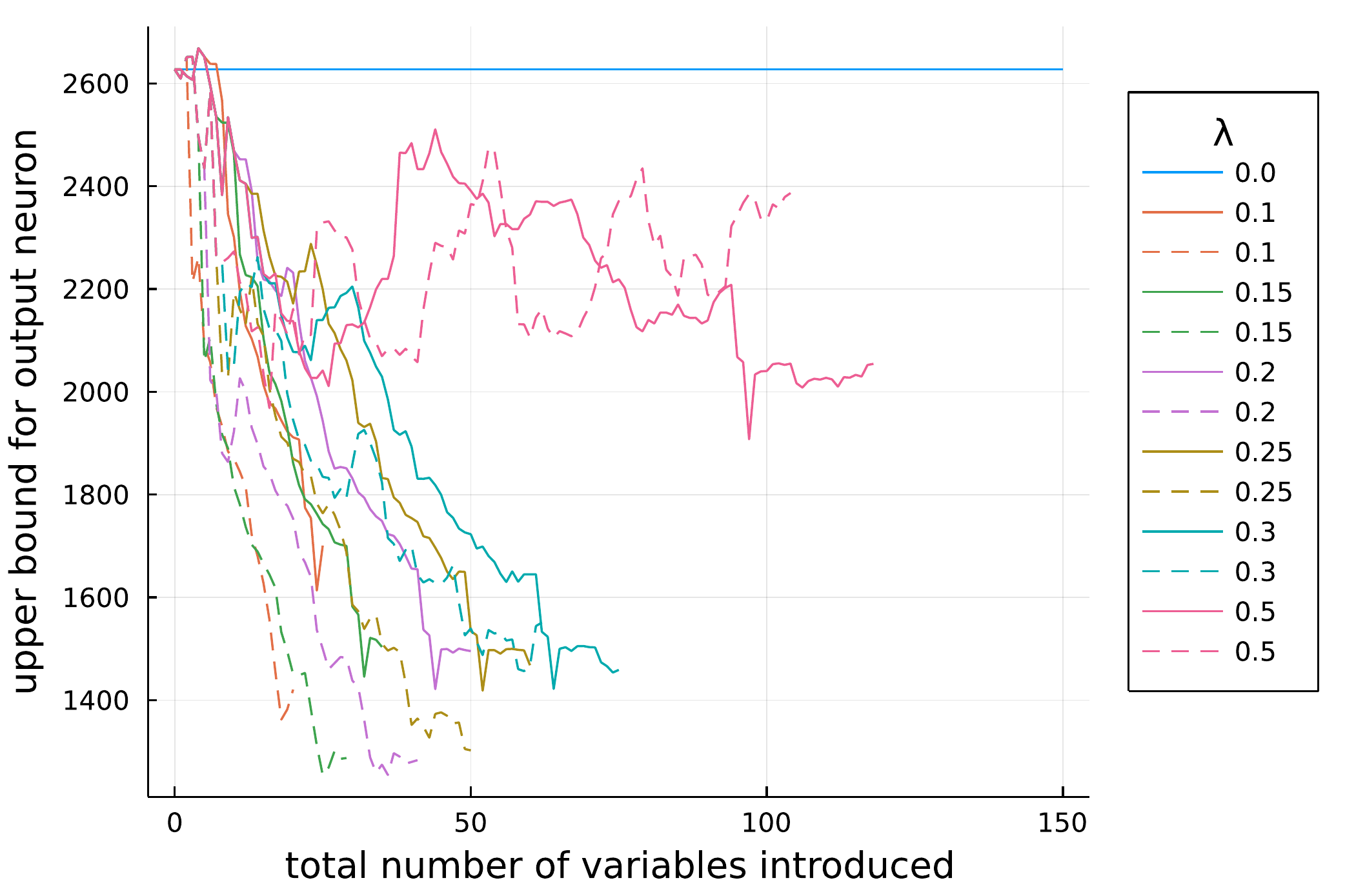}
	\end{subfigure}
	\caption{Upper bounds on the first output of \texttt{ACAS-Xu\_1\_1} for the input space of property $\phi_1$ obtained from a single symbolic forward pass with varying total number of introduced variables. Left: Introducing variables as early as possible for different parameters $\lambda$. No variables ($\lambda = 0$) and the approach of \cite{Paulsen20} ($\lambda = 1$) are shown as solid lines. Right: For different parameters $\lambda$ the effect of introducing a number of variables as early as possible is shown as solid lines, the effect of prioritizing neurons with large concrete ranges and ignoring fixed-zero neurons is shown as dashed lines.}
	\label{fig:var_frac}
\end{figure*}

\subsection{Branch and Bound}
\label{ssec:bab}

Although symbolic interval propagation provides a relatively cheap way of obtaining bounds on the maximal violation $v^*$ in Equation~\ref{eq:nn-opt}, these bounds are not necessarily sufficient to prove the corresponding property. 
A branch and bound algorithm (such as the one presented by Bunel et al. \cite{Bunel20}) uses a \texttt{split} function to repeatedly divide the current problem into smaller sub-problems on which more accurate bounds on the maximal violation can be computed.
A split either divides the input domain into subdomains or an unfixed ReLU activation unit into different phases \cite{Bunel20}.

Different choices for splitting as well as computation of the lower and upper bounds affect the performance of the branch and bound algorithm.
For example \textsc{ZoPE} \cite{Strong21} uses zonotope propagation to obtain upper bounds on maximization problems over the outputs of NNs and samples the center $\vec{x}_{mid}$ of the input hyperrectangle $\mathcal{H}$ to obtain a lower bound $\nnfun{N}(\vec{x}_{mid})$ on the maximal achievable value.
For the split operation, Strong et al. \cite{Strong21} chose to bisect the input $x_i$ with the largest range $[\lo{x_i}, \up{x_i}]$ into two halves.

\section{Our Approach}
\label{sec:approach}

Our approach is based on a combination and thorough refinement of previously presented methods. 
We give a brief overview, before presenting the details.
In order to reduce the effect of the dependency issue, we utilize symbolic interval propagation with fresh variables \cite{Paulsen20} and improve its efficacy by finding heuristics for the amount of variables to introduce per layer and which neurons to prioritize.
These optimizations are described in Sec.~\ref{ssec:fresh_vars}. 

To keep the overapproximations at unstable neurons as small as possible, we utilize the separate relaxations introduced by Wang et al.~\cite{Wang18b} and combine them with the adaptive $\relu$ relaxations published in \cite{Singh19,Zhang18} and propagate the symbolic bounds through 
\begin{align}
	\up{\relu}(x) &= \frac{u_u}{u_u - u_l} x - \frac{u_u u_l}{u_u - u_l} \\
	\lo{\relu}(x) &= \begin{cases}
						0, ~ l_u < |l_l| \\
						x, ~ l_u \geq |l_l|
					\end{cases} \text{ ,}
					\label{eq:dpneurifyfv-relax}
\end{align}
where $[l_l, l_u]$ and $[u_l, u_u]$ are concrete intervals enclosing the values
of $lb(\vec{x})$ and $ub(\vec{x})$ respectively.

If the satisfiability or unsatisfiability of a property cannot be established from a single symbolic forward pass, we make use of the branch and bound implementation developed by Strong et al. \cite{Strong21} based on strong branching and splitting on the input domain.
Although the result is not a complete verifier for $\relu$ networks, it rapidly reduces the obtained bounds for low-dimensional input spaces.
In order to increase the scalability of the branch and bound approach, we devise a new splitting heuristic targeted at reducing overapproximation of unstable neurons, presented in Sec.~\ref{ssec:heuristic}.
In Sec.~\ref{ssec:optimizations}, we present two further optimizations.
The first one ensuring monotonically tighter bounds for smaller sub-problems,
the second one increasing the efficiency of finding counterexamples or good candidate solutions for optimization problems.

\subsection{New Heuristics for Introducing Fresh Variables}
\label{ssec:fresh_vars}

Paulsen et al. \cite{Paulsen20} showed with their tool \textsc{NeuroDiff} that introducing fresh variables can significantly improve performance. 
But they also note that introducing a fresh variable removes all occurrences of previous variables in the symbolic bounds with just the fresh variable.
Therefore introducing a fresh variable diminishes the chances of earlier variables to cancel.
Furthermore, each fresh variable increases the size of the symbolic bounds and consequently makes propagation through the network more expensive.
While we decide to treat the maximum total number of fresh variables as a hyperparameter, Paulsen et al. \cite{Paulsen20} limit the number of fresh variables to
$n_{\text{var}} = \sum_{l=1}^L \sfrac{1}{l} ~ N_l$, 
where $N_l$ is the number of unstable neurons in layer $l$.
To mitigate the first problem in particular, they introduce these $n_{\text{var}}$ variables for the earliest possible unstable neurons in the network.

However, this heuristic does not perform well, when a whole layer of the network consists of only unstable neurons (or fixed-zero and unstable neurons).
In this extreme case all references to earlier variables are removed from the symbolic bounds.
Motivated by this example, we introduce a parameter $\lambda \in [0, 1]$ to bound the number of fresh variables introduced in layer $l$ by
\begin{equation}
	n_{\text{var}_l} = \lambda \cdot n_l \text{ ,}
	\label{eq:var_frac}
\end{equation}
where $n_l$ is the number of neurons in that layer.
If $\lambda$ is low, we introduce fewer variables per layer, but existing variables in earlier layers (including the symbolic input variables) have a higher probability of cancelling out later on.
The effect of this optimization, when we perform a symbolic forward pass for the \texttt{ACAS-Xu\_1\_1} network with increasing maximal number of fresh variables and different values of $\lambda$, can be seen on the left of Fig.~\ref{fig:var_frac}.
The heuristic of \cite{Paulsen20} ($\lambda = 1$) initially improves upon the upper bound obtained from symbolic forward propagation without additional variables, but when too many variables are added, performance becomes significantly worse, as the new variables interfere with each others' cancellation.
For smaller values of $\lambda$, there is no such error-explosion effect, even when a large number of fresh variables is introduced.

While the parameter $\lambda$ controls the maximal fraction of newly introduced variables in a layer, it does not determine \emph{which} neurons on a layer are selected. A further improvement of our heuristic thus selects the neurons on each level, for which new variables are introduced. 

The idea is as follows:
The main advantage of introducing fresh variables is that they have the opportunity to cancel out in later stages of the symbolic forward pass, such that either their symbolic lower or their symbolic upper bound -- never (implicitly) both -- are used in the calculation of concrete bounds for later neurons.
Implicitly using both symbolic bounds of a neuron whose symbolic bounds are far apart likely creates worse bounds than using both symbolic bounds of a neuron whose symbolic lower and upper bound only minimally differ.
Approximating this behaviour, we prioritize introduction of fresh variables for unstable neurons with large ranges between \textit{concrete} lower and upper bounds.

Additionally, we introduce two minor optimizations.
First, instead of directly using the number of neurons in layer $l$ as $n_l$ in Equation~\ref{eq:var_frac}, we choose $n_l$ to be the number of fixed-active or unstable neurons, excluding fixed-inactive neurons,
as their symbolic bounds are all zero.
As our second improvement, we prohibit the introduction of fresh variables in the last hidden layer, as these variables 
have no chance to cancel out before the output layer.
In practice, these two improvements have only a minor effect, as there are few layers with no fixed-active neurons, and when a maximum number of variables is set, there are rarely ones left to be introduced in the last hidden layer.

The overall effect of our fresh-variable-selection-heuristic and the two small optimizations
can be seen on the right of Fig.~\ref{fig:var_frac}, where the solid lines represent the upper bounds obtained from a single symbolic forward pass for property $\phi_1$ through the \texttt{ACAS-Xu\_1\_1} network for some parameters $\lambda$ also shown  on the left of the figure.

\subsection{Splitting Heuristic}
\label{ssec:heuristic}

Introducing fresh variables not only results in tighter bounds obtained from single symbolic forward passes.
By obtaining better bounds, it can also reduce the number of passes required by a branch and bound algorithm to prove a property, often dramatically, thus compensating the slightly higher computation time.

Performance of the branch and bound algorithm can be further improved by better splitting decisions.
Existing input splitting methods have focused on splitting the input with the largest range between its lower and upper bound \cite{Strong21} while others have used gradient information to bisect the input $x_i$ with the largest smear value  \cite{Wang18a}.
Although the gradient captures information about the importance of an input variable on the output of the network, neither the gradient, nor the width of the input interval are a good estimate for the amount of overapproximation-error an input variable is responsible for.
More recently, Henriksen and Lomuscio \cite{Henriksen21} reported significant improvements in verification times, utilizing the error terms in error-based symbolic interval propagation to take interval overapproximation into account.
However, their heuristic is restricted to parallel relaxations for the activation functions.

We therefore adapt their approach and make use of the direct effect of the input variables on the bounds of intermediate neurons.
Given inputs $\vec{x}$ with $x_i \in [\lo{x}_i, \up{x_i}]$ and a neuron $n_j$ with symbolic lower and upper bounds $L_j(\vec{x}) = l_{0j} + \sum_i l_{ij} x_i$ and $U_j(\vec{x}) = u_{0j} + \sum_i u_{ij} x_i$ and concrete bounds $[\lo{L}_j, \up{U}_j]$, 
we can compute the effect of bisecting the range of input $x_i$ into two intervals $[\lo{x}_i, \hat{x}_i]$ and $[\hat{x}_i, \up{x}_i]$, where $\hat{x}_i = \sfrac{1}{2}(\lo{x}_i + \up{x_i})$, on the concretization of $[L_j(\vec{x}), U_j(\vec{x})]$. 
After bisection, we obtain new concrete bounds $\lo{L}_j'(\vec{x})$ and $\up{U}_j'(\vec{x})$:
\begin{align*}
	\lo{L}_j'(\vec{x}) = \begin{cases}
			\lo{L}_j(\vec{x}), & l_{ij} \geq 0, x_i \in [\lo{x}_i, \hat{x}_i]\\
			\lo{L}_j(\vec{x}) - \frac{l_{ij}}{2} (\up{x_i} - \lo{x}_i), & l_{ij} \leq 0, x_i \in [\lo{x}_i, \hat{x}_i]\\
			\lo{L}_j(\vec{x}) + \frac{l_{ij}}{2} (\up{x_i} - \lo{x}_i), & l_{ij} \geq 0, x_i \in [\hat{x}_i, \up{x}_i]\\
			\lo{L}_j(\vec{x}), & l_{ij} \leq 0, x_i \in [\hat{x}_i, \up{x}_i]
		\end{cases}
\end{align*}
\begin{align*}
	\up{U}_j'(\vec{x}) = \begin{cases}
			\up{U}_j(\vec{x}) - \frac{u_{ij}}{2} (\up{x_i} - \lo{x}_i), & u_{ij} \geq 0, x_i \in [\lo{x}_i, \hat{x}_i]\\
			\up{U}_j(\vec{x}), & u_{ij} \leq 0, x_i \in [\lo{x}_i, \hat{x}_i]\\
			\up{U}_j(\vec{x}), & u_{ij} \geq 0, x_i \in [\hat{x}_i, \up{x}_i]\\
			\up{U}_j(\vec{x}) + \frac{u_{ij}}{2} (\up{x_i} - \lo{x}_i), & u_{ij} \leq 0, x_i \in [\hat{x}_i, \up{x}_i]
		\end{cases} \end{align*}
In each of these cases, there is at least one branch where at least one of the bounds is improved and the magnitude of the improvement depends on the range $\up{x_i} - \lo{x}_i$ of the input $x_i$ as well as the \emph{magnitude} $\abs{u_{ij}}$ and $\abs{l_{ij}}$ of the coefficients of $x_i$ in the symbolic lower and upper bound.

In order to reduce the internal overapproximation, we want to split on inputs that cause a large improvement of the bounds of many intermediate neurons.
The importance of an input $x_i$ is therefore
determined as
\begin{align}
	s(x_i) =& \sum_{j=1}^k \frac{1}{2}(\up{x}_i - \lo{x}_i) (\abs{l_{ij}} + \abs{u_{ij}}) \enspace,
\end{align}
where the $n_j$ are the $k$ unstable neurons in the network.
The heuristic then selects the input with largest $s(x_i)$ for splitting.

\subsection{Additional Optimizations}
\label{ssec:optimizations}

If we bisect the input of a NN verification problem to obtain two smaller sub-problems, it is immediately clear that the bounds obtained for the sub-problems have to be at least as tight as the bounds obtained on the original problem, as we maximize or minimize over a smaller input set.
Our symbolic interval propagation algorithm, however, does not naturally return tighter bounds for the sub-problems, as the adaptive $\relu$ relaxations proposed by \cite{Singh19,Zhang18} and the heuristics for introducing fresh variables are sources of non-continuity.

If we split on an input, and thus the concrete bounds on some $\relu$ neuron change, the adaptive lower relaxation might change accordingly and while it still minimizes the area of overapproximation at this particular neuron, it might negatively affect the tightness of the bounds of later neurons.

Similarly for our fresh variable introduction heuristic:
If the concrete bounds of the neurons change due to smaller input spaces, we might introduce fresh variables for different neurons in the sub-problem than in the original problem.
The variables for these neurons might cancel less favourably and thus lead to looser output bounds.

In order to eliminate this behaviour of not monotonically improving bounds, we can store the output bounds of the original problem and only update the bounds in the sub-problem, if those bounds are actually tighter than before.
However, since we minimize of maximize over a smaller subset of the input space not only for the final outputs of the NN, but also for all of its interior neurons, we can proceed in the same manner for these interior neurons and also only update their bounds if they are improved as well.

While the previous optimizations were all concerned with improving the \emph{upper bound} on the maximum value of the optimization problem \ref{eq:nn-opt}, we also propose a heuristic to obtain good candidates for maximizing inputs or counterexamples $\vec{x}$, whose value $\nnfun{N}(\vec{x})$ serves as a \emph{lower bound} to the maximal attainable value of the NN.
To generate candidates for $\vec{x}$, \textsc{ZoPE} \cite{Strong21} periodically samples the center point of the input hyperrectangle of the current sub-problem.
\textsc{Neurify} \cite{Wang18b} on the other hand, uses the input vector that maximizes a linear program containing the symbolic output bounds of the NN and constraints fixing the state of certain intermediate $\relu$ neurons.
The main idea behind this, is that the resulting linear program is a good approximation of the underlying NN, if there is low overapproximation.
In input-splitting however, no additional linear constraints arise from fixing the state of $\relu$ neurons.
Therefore the linear program reduces to just the symbolic interval enclosing the NN's outputs.
We thus make the same assumption that these symbolic intervals $[lb(\vec{x}), ub(\vec{x})]$ are a good approximation of the true outputs of the NN, and use the input that maximizes the symbolic upper bound $ub(\vec{x})$ as candidate for a maximizing input. This operation comes without any additional computational cost, as the calculation of the concrete upper bound $\up{\vec{y}}$ of the output $\nnfun{N}(\vec{x})$ of the NN already depends on setting the corresponding input variables to their appropriate value.

\section{Evaluation}
\label{sec:Evaluation}
The implementation in Julia of our
approach\footnote{Code available at \url{https://github.com/phK3/DPNeurifyFV.jl}}
is called \textsc{DPNeurifyFV} (for \textsc{Neurify} \cite{Wang18b}  with fresh variables and the $\relu$ relaxations used in \textsc{DeepPoly} \cite{Singh19}). It builds upon \texttt{NeuralVerification.jl}\footnote{Code available at \url{https://github.com/sisl/NeuralVerification.jl}.} \cite{Liu21} and \textsc{ZoPE}\footnote{Code available at \url{https://github.com/sisl/NeuralPriorityOptimizer.jl}.} \cite{Strong21}.

In our evaluation, we want to investigate, how \textsc{DPNeurifyFV} compares to state of the art tools in NN verification. Additionally, we want to gain further insights into the effect of our proposed optimizations.
All experiments were run on a single thread and were conducted on a desktop computer running a 64 bit version of Ubuntu 20.04.1 with an Intel Core i7-2600 3.4 GHz processor, which has 8 cores, and 8GB of RAM.

\subsection{The ACAS Xu Benchmark.}

To demonstrate the effectiveness of our approach, we use the ACAS Xu neural network verification benchmark set published by Katz et al. \cite{Katz2017}.
The benchmark consists of NNs that were trained to provide navigation advisories on board of aeroplanes as part of the Airborne Collision Avoidance System for unmanned aircraft. 
Each of the $45$ NNs has $5$ input neurons taking in parameters of the own aircraft and aircraft nearby, $6$ fully connected layers of $50$ neurons each, and $5$ output neurons representing scores for the different navigation advisories.
We consider properties $\phi_1, \phi_2, \phi_3$ and $\phi_4$ of the benchmark set, which are to be checked for each of the $45$ networks.
All of the properties define a hyperrectangle of valid values for the input space and an appropriate property on the output space of the NN.
Property $\phi_1$ asks for an upper bound on the first output of the network and can therefore be modelled as maximizing the violation of a single linear constraint, while properties $\phi_2, \phi_3$ and $\phi_4$ require that certain outputs never have the minimal or maximal score among all outputs of the network.
As the constraints $o_i \geq o_j, ~\forall i \neq j$, where the $o_k$ are the outputs of the NN, represent a convex polytope, we want to ensure that the output set of the NN never intersects this polytope.
Therefore, these latter properties can be modelled by minimizing the violation of the associated constraints of the polytope.

\begin{figure}[H]
	\includegraphics[width=\linewidth]{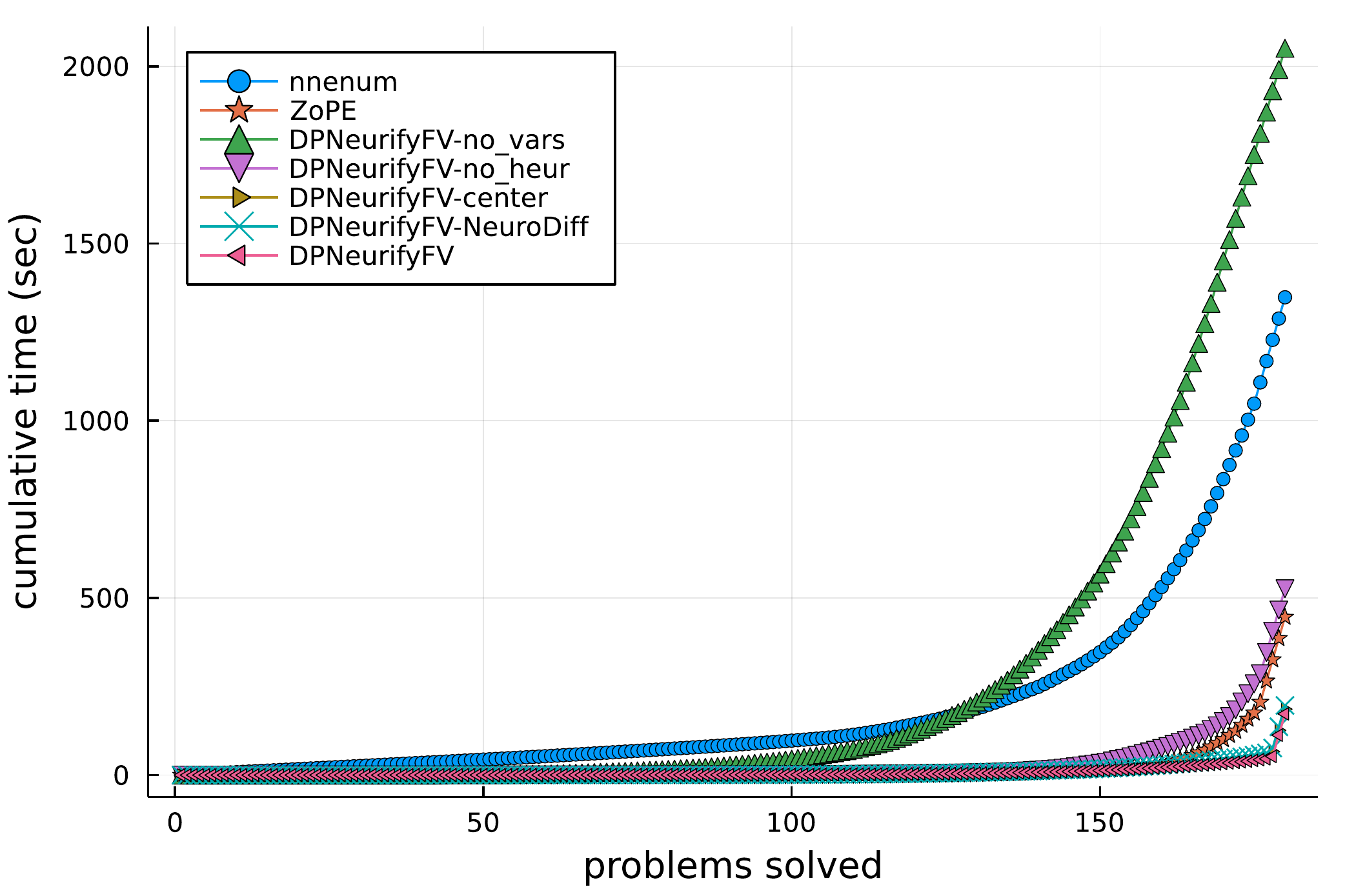}
	\caption{Overall solving time for properties $\phi_1, \phi_2, \phi_3, \phi_4$ over all networks of the \texttt{ACAS-Xu} family for \textsc{ZoPE}, \textsc{NNENUM}, our tool \textsc{DPNeurifyFV} and ablations of our tool without introducing fresh variables, without utilizing our splitting heuristic, sampling the center point instead of the maximizer/minimizer of the symbolic bound to obtain a candidate solution and using the \textsc{NeuroDiff} heuristic for fresh variable introduction.}
	\label{fig:cactus_all}
\end{figure}

\subsection{Verification.}

We run our tool \textsc{DPNeurifyFV} on all the aforementioned ACAS Xu
verification tasks,
and compare its performance to \textsc{ZoPE} \cite{Strong21}, a solver for optimization problems over the output of NNs and \textsc{NNENUM}\footnote{Code available at \url{https://github.com/stanleybak/nnenum}.} \cite{Bak20}, the NN verifier that scored highest amongst all participants at \textsc{VNN-COMP 2021} \cite{Bak21} on the ACAS Xu benchmark set.
We configured \textsc{DPNeurifyFV} to introduce at most $20$ fresh variables and to introduce fresh variables for a maximal fraction $\lambda = \sfrac{1}{2}$ of a layer's non-fixed zero neurons.
For \textsc{ZoPE}, we used the setting the authors published in the appendix of their paper \cite{Strong21}, while we used the standard parameters for the configuration of \textsc{NNENUM} \cite{Bak20}, only changing the number of processes to one.
For all methods, we set a timeout of $60$ seconds per instance.
The overall result is shown in Table~\ref{tab:acas-verify} and displayed as a cactus plot in Fig.~\ref{fig:cactus_all}. 
The results indicate that \textsc{DPNeurifyFV} is significantly faster than \textsc{NNENUM}. While \textsc{NNENUM} finishes verification of $175$ of the $180$ tasks in $1,048.2$ seconds, our tool successfully verifies $178$ of the properties in just $52.73$ seconds -- a $19 \times$ improvement.
In comparison to \textsc{ZoPE}'s runtime of $205.9$ seconds for the verification of $176$ properties, our approach is still considerably faster by a factor of $4$.
when minimizing the violation of the associated constraints.

To investigate the effect of our different optimizations, we compare the full implementation of our approach to variants without certain optimizations.
We refer to the version of \textsc{DPNeurifyFV} without the introduction of fresh variables as \textsc{DPNeurifyFV-no\_vars}, without the usage of our splitting heuristic as \textsc{DPNeurifyFV-no\_heur}, to the version without using the maximizer/minimizer of the symbolic output bounds to find good solution candidates as \textsc{DPNeurifyFV-center}, as it just samples the center point of the input hyperrectangle of the current sub-problem and to the version that uses the heuristic of \textsc{NeuroDiff} \cite{Paulsen20} for introduction of fresh variables as \textsc{DPNeurifyFV-NeuroDiff}. 

The effect of introducing fresh variables is the largest improvement by far, as can be seen from the performance of  \textsc{DPNeurify-no\_vars}, which is significantly slower than \textsc{NNENUM} on both satisfiable as well as unsatisfiable instances.
Due to the dependency issue, the obtained symbolic bounds do neither tightly approximate the true optimization target, nor the input bounds to the intermediate neurons, thus leading to large overapproximations.

\begin{table*}[h]
\centering
\caption{Number of solved instances and cumulative verification times for properties $\phi_1, \phi_2, \phi_3$ and $\phi_4$ of the ACAS Xu verification benchmark (180 instances total).}
\vspace{5pt}
\begin{tabular}{l||cr|cr|c}
\toprule
 Approach & \texttt{SAT} & (sec) & \texttt{UNSAT} & (sec) & \texttt{INCONCLUSIVE}\\ 
\midrule
\textsc{NNENUM} \cite{Bak20} & $44$ & $114.09$ & $131$ & $934.11$ & $5$\\  \hline 
\textsc{ZoPE} \cite{Strong21} & $44$ & $81.51$ & $132$ & $124.39$ & $4$\\ \hline 
\textsc{DPNeurifyFV}-no\_vars & $\textbf{45}$ & $137.33$ & $124$ & $1,253.53$ & $11$\\ 
\textsc{DPNeurifyFV}-no\_heur & $\textbf{45}$ & $100.17$ & $132$ & $245.58$ & $3$\\ 
\textsc{DPNeurifyFV}-center & $\textbf{45}$ & $25.78$ & $\textbf{133}$ & $40.48$ & $\textbf{2}$\\ 
\textsc{DPNeurifyFV-NeuroDiff} & $\textbf{45}$ & $20.40$ & $\textbf{133}$ & $56.02$ & $\textbf{2}$\\

\textsc{DPNeurifyFV} & $\textbf{45}$ & $\textbf{12.82}$ & $\textbf{133}$ & $\textbf{39.91}$ & $\textbf{2}$\\  \bottomrule
\end{tabular}
\label{tab:acas-verify}
\end{table*}

\section{Conclusion and Future Work}
\label{sec:conclusion}

In this paper, we presented \textsc{DPNeurifyFV}, an algorithm for verification of $\relu$ NNs with low-dimensional input space.
The approach is based on input splitting and combining several methods for symbolic interval propagation.
We added important further optimizations encompassing new heuristics for introducing fresh variables, a novel splitting heuristic that can be used for a wide set of $\relu$ relaxations and further smaller improvements of the branch-and-bound search.

We evaluated \textsc{DPNeurifyFV} on the ACAS Xu benchmark set and compared it to the state-of-the-art verification tools \textsc{ZoPE} and \textsc{NNENUM}, achieving considerable improvements in runtime.
We also empirically evaluated the effect of our different improvements, showing that introducing new variables is essential, but often only in combination with further optimizations.
We also empirically evaluated the effect of our different improvements, showing that
introducing new variables is essential, but often only in combination with other
optimizations.

Future work might include an adaptation of our method to the equivalence verification
problem for NNs \cite{KleineBuening20,Teuber21} as well as an evaluation
of our approach on further benchmarks.

\section*{Acknowledgements}

The authors would like to thank the Ministry of Science,
Research and Arts of the Federal State of Baden-W\"{u}rttemberg, Germany
for financial support within the InnovationCampus Future Mobility.

\bibliography{PaperFreshVars}
\bibliographystyle{wfvml2022}

\end{document}